\documentclass[conference]{IEEEtran}
\usepackage{cite}
\usepackage{amsmath,amssymb,amsfonts}
\usepackage{graphicx}
\usepackage{textcomp}
\usepackage{booktabs}
\usepackage{array}
\usepackage{hyperref}

\begin{document}

\title{Robust OT-Guided Generative Residual Domain Adaptation for Bike-Sharing Demand Prediction under Temporal Domain Shift}

\author{\IEEEauthorblockN{Yiming Ma}
\IEEEauthorblockA{Department of Statistics and Finance, School of Management,\\
University of Science and Technology of China\\
Email: mayiming@mail.ustc.edu.cn}}

\maketitle

\begin{abstract}
Bike-sharing models trained on historical station-hour data may degrade when deployed in later years because travel patterns change over time. This paper studies March Citi Bike demand prediction from 2021 to 2026 as a temporal domain adaptation problem and proposes Gen-ROTDA, a robust optimal transport-guided residual domain adaptation framework. The method fits a target-domain station-time anchor with a small labeled target subset, transfers residual rather than raw demand, applies a deterministic label-preserving residual feature generator, and trims high-cost transport matches before training the final residual predictor. Experiments compare Gen-ROTDA with anchor-only, source-only, target-only, fine-tuning, MMD adaptation, Sinkhorn OTDA, ROTDA, and Gen-OTDA. Gen-ROTDA achieves the lowest MAE on the main 2025 to 2026 task and is the best OT-family method on average across multi-year tasks, although fine-tuning and MMD adaptation remain strong overall baselines. Under abnormal target-unlabeled records, Gen-ROTDA is much more stable than non-robust OT variants, suggesting that robust transport is useful for noisy temporal transfer in bike-sharing demand prediction.
\end{abstract}


\section{Introduction}

Accurate bike-sharing demand prediction helps operators allocate bikes, rebalance stations, and maintain service quality~\cite{omahony2015,vogel2011}. In practice, a model trained on historical station-hour observations is often deployed in a later year. This creates temporal domain shift because station popularity, commuting behavior, rider composition, and operational patterns may change over time. Such distribution changes are central to transfer learning and domain adaptation~\cite{bendavid2010,pan2010}, and they can make direct source-year prediction unreliable.

Cross-year station-hour prediction is especially challenging because demand is sparse, skewed, and noisy. Many station-hours have low demand, while a small number of stations and peak hours dominate prediction errors. The target feature pool may also contain abnormal records caused by unusual operations, events, weather, or data-quality issues. A useful adaptation method should therefore align source and target feature distributions while remaining stable under abnormal target records.

Optimal transport (OT) provides a natural way to align empirical distributions for domain adaptation~\cite{courty2017jd,courty2017ot}. Entropic OT, commonly computed by Sinkhorn iterations, improves numerical stability and scalability~\cite{cuturi2013,peyre2019}. However, standard OT may still be affected by poorly matched or abnormal samples because high-cost matches can influence the transport plan. Robust OT addresses this issue by trimming or down-weighting abnormal transport mass~\cite{ma2025,mukherjee2021}.

This paper proposes Gen-ROTDA, a robust OT-guided residual domain adaptation method for bike-sharing demand prediction. The method uses an anchor-residual design: a target-domain station-time anchor captures stable structure, and domain adaptation is applied only to the residual demand component. Gen-ROTDA then combines a deterministic label-preserving residual feature generator with robust OT trimming. The generator is not a GAN or a stochastic generative model; it is a residual feature transformation network that moves source residual features toward the target domain while preserving demand-relevant information, following the broader idea of transferable representation learning through distribution alignment~\cite{ganin2016,gretton2012,long2015,sun2016}.

The contributions are threefold. First, we formulate cross-year Citi Bike station-hour prediction as a residual temporal domain adaptation problem. Second, we propose Gen-ROTDA, which combines label-preserving residual feature transformation and robust OT alignment. Third, we evaluate the method on March Citi Bike data from 2021 to 2026 using main-task, multi-year, robustness, ablation, and visualization experiments.

\section{Data and Prediction Task}

\subsection{Citi Bike Station-Hour Data}

The experiments use Citi Bike trip records aggregated to station-hour observations from the public Citi Bike system data~\cite{citibike2026}. For station $s$ and hour $t$, the target variable is
\begin{equation}
Y_{s,t} = \text{number of trips starting from station } s \text{ during hour } t.
\label{eq:demand}
\end{equation}

Each observation is represented as $(X_{s,t}, Y_{s,t})$. The processed data cover March of each year from 2021 to 2026. Using the same calendar month reduces seasonal confounding and focuses the experiments on cross-year temporal shift. The processed files contain station identifiers, station coordinates, calendar fields, lagged demand, rolling demand summaries, rider and bike-type counts, and ratio variables.

The main residual transfer experiments use a compact feature split. The variables are classified into three groups. The anchor features include spatial and calendar variables, namely \texttt{start\_lat}, \texttt{start\_lng}, \texttt{hour\_sin}, \texttt{hour\_cos}, \texttt{dow\_sin}, \texttt{dow\_cos}, and \texttt{is\_weekend}. The transfer features include demand-history variables, namely \texttt{lag\_1h}, \texttt{lag\_24h}, \texttt{rolling\_24h\_mean}, and \texttt{rolling\_168h\_mean}. The target variable is demand, which represents the station-hour bike demand to be predicted.

The transfer features are transformed by log1p and standardized before domain adaptation. The lag and rolling features are computed causally from past observations available before the prediction time. The prediction setting is therefore short-horizon station-hour forecasting with historical demand information available, not a setting in which future target labels are used to construct features.

\subsection{Temporal Domain Adaptation Setting}

Let
\begin{equation}
\mathcal{D}_s = \{(X_i^s, Y_i^s)\}_{i=1}^{n_s}
\label{eq:source}
\end{equation}
denote labeled source-year samples, and let
\begin{equation}
\mathcal{D}_t = \{(X_j^t, Y_j^t)\}_{j=1}^{n_t}
\label{eq:target}
\end{equation}
denote target-year samples. During adaptation, target features are available, and a small labeled target subset is used for anchor fitting and residual calibration. The held-out target labels are used only for final evaluation. All compared methods use the same labeled target subset and the same target-unlabeled feature pool. The goal is to predict held-out target demand under
\begin{equation}
P_s(X,Y) \neq P_t(X,Y).
\label{eq:shift}
\end{equation}

The main task is 2025 to 2026 transfer. Additional tasks include adjacent-year transfers from 2021 to 2022 through 2024 to 2025 and two-year transfers from 2021 to 2023 through 2024 to 2026.

\section{Method}

\begin{figure}[t]
\centering
\includegraphics[width=0.82\columnwidth]{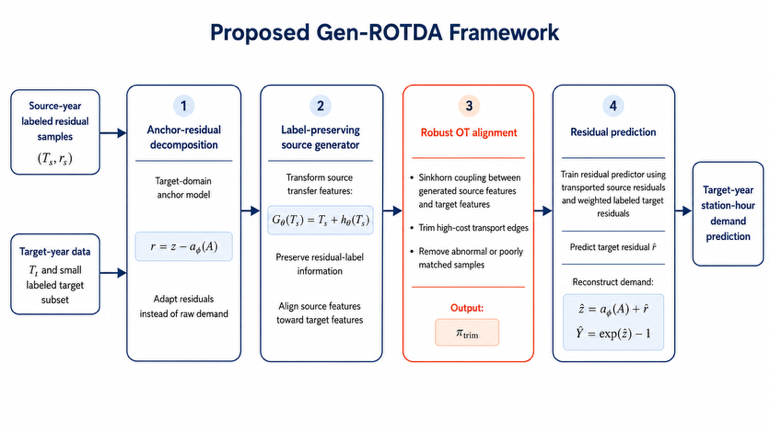}
\caption{Overview of the proposed Gen-ROTDA framework for robust residual domain adaptation in station-hour demand prediction.}
\label{fig:framework}
\end{figure}

Figure~\ref{fig:framework} illustrates the proposed Gen-ROTDA framework for target-year station-hour demand prediction. The method first decomposes demand into an anchor component and a residual component. It then uses a label-preserving generator to adapt source residual features toward the target domain, followed by robust optimal transport alignment to remove poorly matched samples. Finally, a residual predictor is trained using the transported source residuals and labeled target residuals, and the final demand prediction is obtained by combining the anchor prediction with the predicted residual.

\subsection{Anchor-Residual Decomposition}

Directly transferring raw demand can be unstable because demand contains both stable station-time effects and year-specific residual dynamics. We therefore use an anchor-residual decomposition on the log-demand scale. Let
\begin{equation}
z = \log(1 + Y).
\label{eq:logdemand}
\end{equation}

A target-domain anchor model $a_{\phi}(A)$ is trained on labeled target samples using anchor features $A$. The residual is
\begin{equation}
r = z - a_{\phi}(A).
\label{eq:residual}
\end{equation}

The final prediction is
\begin{equation}
\widehat{z} = a_{\phi}(A) + \widehat{r}, \qquad \widehat{Y} = \exp(\widehat{z}) - 1.
\label{eq:finalpred}
\end{equation}

Domain adaptation is applied only to residual prediction. This design allows the small labeled target subset to calibrate stable station-time structure, while the source year contributes transferable residual demand patterns.

\subsection{Label-Preserving Residual Feature Generator}

Let $T$ denote the standardized transfer feature vector. Gen-OTDA and Gen-ROTDA use a deterministic residual feature generator
\begin{equation}
G_{\theta}(T) = T + h_{\theta}(T),
\label{eq:generator}
\end{equation}
where $h_{\theta}$ is a small multilayer perceptron. The residual form discourages excessive feature movement and keeps the transformed source features close to their original representation.

The generator is trained to make generated source features resemble target features while preserving residual-label information, combining distribution matching ideas from MMD-based adaptation with label-aware transport intuition~\cite{courty2017jd,courty2017ot,gretton2012,long2015}. The objective is
\begin{multline}
\mathcal{L}_G = \mathcal{L}_{\mathrm{align}}(G_{\theta}(T^s), T^t) + \lambda_{\mathrm{id}} \| G_{\theta}(T^s) - T^s \|_2^2 \\
+ \lambda_{\mathrm{lp}} \mathcal{L}_{\mathrm{label}} + \lambda_{\mathrm{sup}} \mathcal{L}_{\mathrm{target}}.
\label{eq:genloss}
\end{multline}

Here $\mathcal{L}_{\mathrm{align}}$ is an MMD-style distributional alignment loss~\cite{gretton2012,long2015}, $\mathcal{L}_{\mathrm{label}}$ trains a source residual prediction head on generated source features, and $\mathcal{L}_{\mathrm{target}}$ uses labeled target residuals when available. This objective makes the transformed source features more target-like without ignoring the residual demand signal.

\subsection{OT and Robust OT Alignment}

After source features are generated, OT-based methods align source and target transfer-feature distributions~\cite{courty2017jd,courty2017ot}. For generated source features $\widetilde{T}_i^s = G_{\theta}(T_i^s)$ and target features $T_j^t$, the squared Euclidean cost is
\begin{equation}
C_{ij} = \| \widetilde{T}_i^s - T_j^t \|^2.
\label{eq:cost}
\end{equation}

Sinkhorn OTDA computes an entropic transport coupling~\cite{cuturi2013,peyre2019}
\begin{equation}
\pi^{\star} = \arg\min_{\pi \in \Pi(a,b)} \langle \pi, C \rangle + \varepsilon \sum_{i,j} \pi_{ij} (\log \pi_{ij} - 1).
\label{eq:sinkhorn}
\end{equation}

Gen-ROTDA adds a robust trimming step motivated by outlier-robust OT~\cite{ma2025,mukherjee2021}. Coupling entries are sorted by transport cost, and high-cost entries are removed until the retained coupling mass reaches the specified $\texttt{keep\_mass}$. The retained coupling is renormalized and used for barycentric transport:
\begin{equation}
\overline{T}_i^s = \frac{\sum_j \pi_{ij}^{\mathrm{trim}} T_j^t}{\sum_j \pi_{ij}^{\mathrm{trim}}}.
\label{eq:barycentric}
\end{equation}

If trimming removes all retained mass from a source row, the implementation keeps the generated source feature for that row as a fallback. This avoids an undefined barycentric projection. The final residual predictor is trained on transported source residuals and labeled target residuals, with labeled target residuals up-weighted by $\texttt{target\_weight}$.

\subsection{Compared Methods}

The comparison includes nine methods. Anchor-only uses only the target-domain station-time anchor model, while Source-only trains a residual predictor on source data and directly applies it to the target year. Target-only trains the residual predictor using only labeled target samples, and Fine-tuning pools source residuals with weighted labeled target residuals. MMD adaptation performs source adaptation or reweighting based on MMD. OTDA uses ordinary optimal transport alignment without a generator or robust trimming, while Sinkhorn OTDA uses entropic optimal transport alignment without robust trimming. ROTDA applies robust optimal transport trimming without a generator. Gen-OTDA combines the residual feature generator with non-robust optimal transport alignment, whereas Gen-ROTDA combines the residual feature generator with robust optimal transport trimming.

\section{Experimental Setup}

The default experiments use March station-hour observations. The main experiment uses 2025 as the source year and 2026 as the target year. The random seed is fixed at 2026.

The experimental setup uses 1,000 source training samples and 1,000 unlabeled target samples. The maximum number of labeled target samples is 500, and the maximum number of target test samples is 3,000. Seven target labeled days are used. The transfer feature set is \texttt{lag\_only}. The target weight is set to 8, the Sinkhorn regularization scale is 0.1, and the robust retained mass is 0.8. The generator is trained for 200 epochs.

The base prediction model is a random forest regressor with 300 trees and \texttt{min\_samples\_leaf}=3~\cite{breiman2001}.

Prediction performance is evaluated on held-out target station-hour observations. The main metrics are mean absolute error (MAE), root mean squared error (RMSE), and $R^2$:
\begin{equation}
\mathrm{MAE} = \frac{1}{n} \sum_{i=1}^{n} |\widehat{Y}_i - Y_i|,
\label{eq:mae}
\end{equation}
\begin{equation}
\mathrm{RMSE} = \left( \frac{1}{n} \sum_{i=1}^{n} (\widehat{Y}_i - Y_i)^2 \right)^{1/2},
\label{eq:rmse}
\end{equation}
\begin{equation}
R^2 = 1 - \frac{\sum_i (Y_i - \widehat{Y}_i)^2}{\sum_i (Y_i - \bar{Y})^2}.
\label{eq:rsq}
\end{equation}

MAE is the primary metric because station-hour demand is sparse and skewed. RMSE is also reported because it is more sensitive to large errors.

\section{Results}

\subsection{Main 2025 to 2026 Transfer Result}

Table~\ref{tab:main} reports the main 2025 to 2026 transfer result. Gen-ROTDA obtains the lowest MAE, reducing MAE by 2.36\% relative to source-only learning and by 0.09\% relative to Sinkhorn OTDA. The improvement over ROTDA is small, so the main conclusion is not that the generator alone creates a large accuracy gain. Rather, robust transport is the main source of MAE improvement, and the generator gives a modest additional benefit in this task.

\begin{table}[t]
\centering
\caption{Main 2025 to 2026 transfer results.}
\label{tab:main}
\footnotesize
\begin{tabular}{lccccc}
\toprule
Method & MAE & RMSE & $R^2$ & Runtime (s) \\
\midrule
Anchor-only    & 0.9364 & 1.9841 & 0.0536 & 0 \\
Source-only    & 0.8834 & 1.6375 & 0.3553 & 0.1185 \\
Target-only    & 0.8942 & 1.8582 & 0.1699 & 0.1074 \\
Fine-tuning    & 0.8697 & 1.6597 & 0.3377 & 0.1452 \\
MMD adaptation & 0.8697 & 1.6601 & 0.3374 & 0.1858 \\
Sinkhorn OTDA  & 0.8633 & 1.6527 & 0.3434 & 11.8844 \\
ROTDA          & 0.8629 & 1.6557 & 0.3410 & 12.0745 \\
Gen-OTDA       & 0.8679 & 1.6453 & 0.3492 & 9.8388 \\
Gen-ROTDA      & \textbf{0.8625} & 1.6614 & 0.3364 & 9.4047 \\
\bottomrule
\end{tabular}
\end{table}

The RMSE pattern is mixed. Gen-OTDA has the lowest RMSE and highest $R^2$, while Gen-ROTDA has the lowest MAE. This is consistent with the purpose of robust trimming: it is designed to improve stability and average absolute error under noisy matching, not necessarily to minimize large-error-sensitive RMSE.

\subsection{Multi-Year Transfer Comparison}

Table~\ref{tab:multiyear} averages results over four adjacent-year tasks and four two-year tasks. Fine-tuning and MMD adaptation have the best overall average MAE. Gen-ROTDA is therefore not the best method across all baselines in the multi-year average. However, it remains the best OT-family method, slightly improving over Sinkhorn OTDA, ROTDA, and Gen-OTDA.

\begin{table}[t]
\centering
\caption{Average performance across multi-year transfer tasks.}
\label{tab:multiyear}
\small
\begin{tabular}{lcccc}
\toprule
Method & Adjacent-year MAE & Two-year MAE & Overall MAE & Overall RMSE \\
\midrule
Anchor-only    & 0.8809 & 0.8713 & 0.8761 & 1.6254 \\
Source-only    & 0.8457 & 0.8293 & 0.8375 & 1.5029 \\
Target-only    & 0.8535 & 0.8388 & 0.8462 & 1.5312 \\
Fine-tuning    & 0.8422 & 0.8163 & 0.8293 & 1.4864 \\
MMD adaptation & 0.8419 & 0.8156 & \textbf{0.8281} & \textbf{1.4823} \\
Sinkhorn OTDA  & 0.8476 & 0.8249 & 0.8363 & 1.4976 \\
ROTDA          & 0.8475 & 0.8216 & 0.8346 & 1.4883 \\
Gen-OTDA       & 0.8479 & 0.8239 & 0.8359 & 1.4926 \\
Gen-ROTDA      & 0.8458 & 0.8226 & 0.8342 & 1.4883 \\
\bottomrule
\end{tabular}
\end{table}

The identical aggregate values for fine-tuning and MMD adaptation indicate that, under the current residual setup, the MMD adaptation step produces almost the same predictive solution as simple fine-tuning. We therefore interpret this baseline cautiously: it is a strong simple adaptation baseline, but the table does not provide separate evidence that MMD weighting adds benefit beyond fine-tuning in this experiment.

\subsection{Robustness under Abnormal Target Records}

To evaluate robustness, abnormal records are injected into the target-unlabeled feature pool at contamination ratios from 0\% to 20\%. Table~\ref{tab:robustness} reports MAE for Sinkhorn OTDA, Gen-OTDA, and Gen-ROTDA. Gen-ROTDA is the most stable method under high contamination. At 20\% contamination, Sinkhorn OTDA increases from 0.8634 to 0.8810, and Gen-OTDA increases from 0.8678 to 0.8876. In contrast, Gen-ROTDA changes only from 0.8625 to 0.8629.

\begin{table}[t]
\centering
\caption{Robustness results under abnormal target-unlabeled records.}
\label{tab:robustness}
\small
\begin{tabular}{lccc}
\toprule
Abnormal ratio & Sinkhorn OTDA & Gen-OTDA & Gen-ROTDA \\
\midrule
0.00 & 0.8634 & 0.8678 & 0.8625 \\
0.05 & 0.8675 & 0.8694 & 0.8650 \\
0.10 & 0.8771 & 0.8771 & 0.8590 \\
0.15 & 0.8744 & 0.8830 & 0.8583 \\
0.20 & 0.8810 & 0.8876 & \textbf{0.8629} \\
\bottomrule
\end{tabular}
\end{table}

\begin{figure}[t]
\centering
\includegraphics[width=0.82\columnwidth]{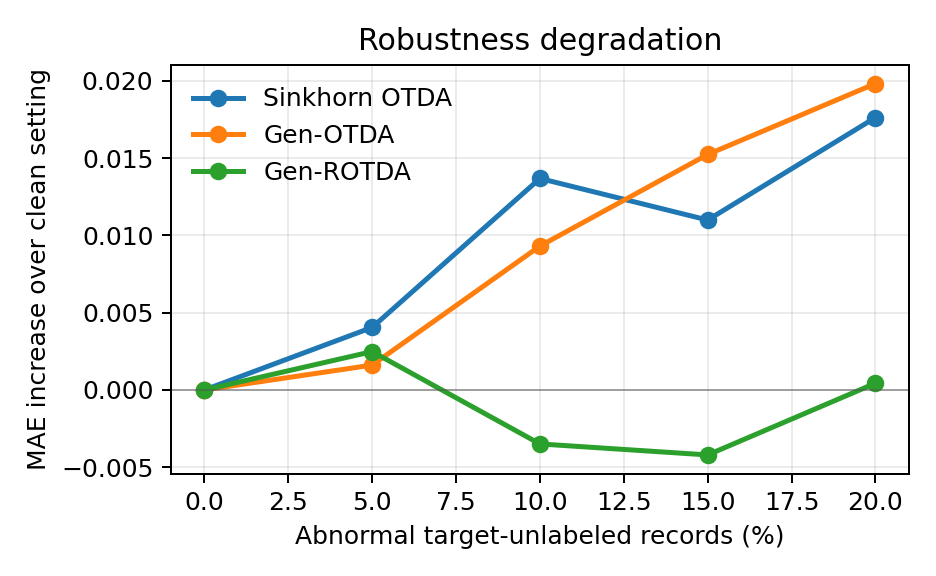}
\caption{Robustness degradation under abnormal target records.}
\label{fig:robustness}
\end{figure}

\subsection{Ablation Study}

Table~\ref{tab:ablation} separates the effects of the generator and robust transport. Adding robust trimming to OTDA improves MAE from 0.8790 to 0.8628. Adding the generator without robust trimming gives Gen-OTDA an MAE of 0.8678. Combining the generator with robust trimming gives the best MAE, 0.8624, but the gain over ROTDA is small.

\begin{table}[t]
\centering
\caption{Ablation study on the 2025 to 2026 task.}
\label{tab:ablation}
\small
\begin{tabular}{lccccc}
\toprule
Method & Generator & Robust OT & MAE & RMSE & $R^2$ \\
\midrule
OTDA     & No  & No  & 0.8790 & 1.6488 & 0.3465 \\
ROTDA    & No  & Yes & 0.8628 & 1.6555 & 0.3412 \\
Gen-OTDA & Yes & No  & 0.8678 & 1.6451 & 0.3494 \\
Gen-ROTDA & Yes & Yes & \textbf{0.8624} & 1.6605 & 0.3372 \\
\bottomrule
\end{tabular}
\end{table}

The ablation confirms that robust OT is the dominant contributor to MAE improvement in the current residual setup. The generator is still useful as a flexible feature transformation, but its empirical contribution is modest unless combined with robust trimming.

\section{Visualization and Interpretation}

\begin{figure}[t]
\centering
\includegraphics[width=0.82\columnwidth]{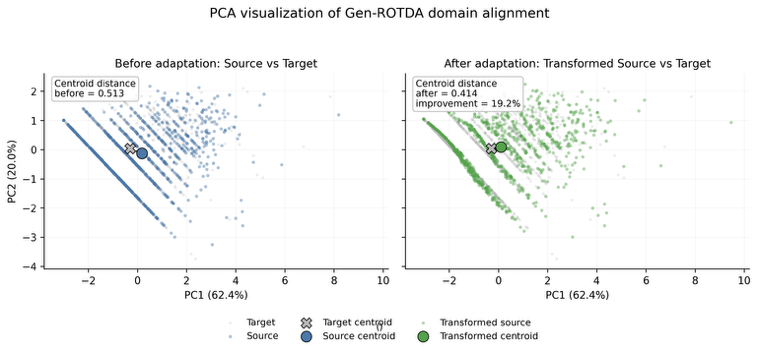}
\caption{PCA domain alignment for 2025 to 2026.}
\label{fig:pca}
\end{figure}

Figure~\ref{fig:pca} visualizes source, target, and generated source feature distributions by PCA for the 2025 to 2026 setting. The PCA plot uses standardized log transfer features. The centroid distance between source and target decreases from 0.5125 to 0.4140 after Gen-ROTDA transformation, corresponding to a 19.22\% reduction. The mean feature displacement is 0.3965 and the median displacement is 0.4115, suggesting a moderate transformation rather than a destructive remapping.

\begin{figure}[t]
\centering
\includegraphics[width=0.82\columnwidth]{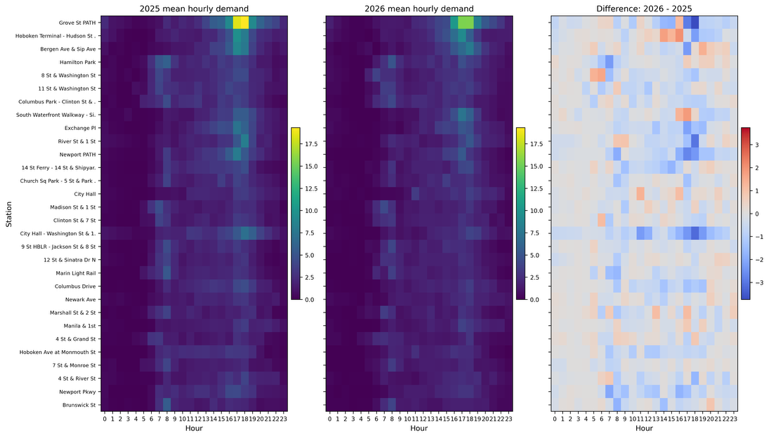}
\caption{Demand heatmap for 2025 versus 2026.}
\label{fig:heatmap}
\end{figure}

Figure~\ref{fig:heatmap} shows station-hour demand shift between March 2025 and March 2026 for representative stations. The summary contains 720 station-hour cells from 30 stations. The mean absolute demand difference is 0.442 trips per station-hour, and the largest absolute difference is 3.742 trips per station-hour. Large negative shifts appear at important commute stations such as Grove St PATH during evening peak hours.

These visualizations support the experimental setting. The feature distributions are shifted but not completely unrelated, making residual domain adaptation plausible. The demand heatmap also shows that some station-hour patterns change substantially across years, which explains why direct source-only transfer is insufficient and why target-aware residual calibration is needed.

\section{Discussion}

The results show three main patterns. First, the anchor-residual formulation is important. Anchor-only prediction is not sufficient, but the anchor captures stable station-time structure and makes residual transfer easier. This also explains why fine-tuning and MMD adaptation are strong baselines: once the stable structure is removed, the remaining residual problem is lower dimensional and can benefit directly from a small labeled target subset.

Second, robust OT is the most reliable source of improvement, consistent with the known sensitivity of standard OT to outliers~\cite{mukherjee2021}. In the main experiment and the ablation study, ROTDA and Gen-ROTDA improve MAE relative to non-robust OT variants. Under abnormal target-unlabeled records, robust trimming is especially important. Gen-ROTDA remains nearly unchanged at 20\% contamination, whereas Sinkhorn OTDA and Gen-OTDA degrade more noticeably.

Third, the generator provides additional flexibility but should not be overstated. The generator helps create target-like source residual features and can improve the final robust OT method slightly, but the multi-year results show that simple fine-tuning and MMD adaptation remain very strong. The current evidence therefore supports Gen-ROTDA mainly as a robust OT-family method rather than a universally best predictor across all baselines.

The method has several practical advantages. It is compatible with standard regression models, uses interpretable lagged demand features, and can be evaluated with conventional forecasting metrics. The robust trimming mechanism is simple and directly targets abnormal transport matches. The main limitations are the number of hyperparameters and the use of March data only. Future work should evaluate more months, include weather and event covariates, and develop adaptive rules for choosing the retained transport mass.

\section{Conclusion}

This paper proposes Gen-ROTDA, a robust OT-guided residual domain adaptation framework for cross-year bike-sharing demand prediction. The method decomposes demand into a target-domain station-time anchor and a transferable residual component, applies a deterministic label-preserving residual feature generator, and uses robust OT trimming before fitting the final residual predictor. Experiments on Citi Bike station-hour data from 2021 to 2026 show that Gen-ROTDA achieves the best MAE on the main 2025 to 2026 task, is the best OT-family method on average across multi-year tasks, and is much more stable under abnormal target records than non-robust OT variants. The findings suggest that robust transport is useful for temporal domain adaptation in noisy urban mobility forecasting, while the benefit of generative residual feature transformation should be interpreted as complementary and task-dependent.

\bibliographystyle{IEEEtran}
\bibliography{references}

\end{document}